\documentclass[a4paper,twoside]{article}

\usepackage{epsfig}
\usepackage{subcaption}
\usepackage{calc}
\usepackage{amssymb}
\usepackage{amstext}
\usepackage{amsmath}
\usepackage{amsthm}
\usepackage{multicol}
\usepackage{pslatex}
\usepackage{apalike}
\usepackage{algorithm2e}
\usepackage[bottom]{footmisc}
\usepackage{SCITEPRESS}     

\usepackage{amsmath,amssymb,amsfonts}
\usepackage{graphicx}
\usepackage{textcomp}
\usepackage{xcolor}
\usepackage{algorithm2e}
\usepackage{subcaption}
\usepackage{url}
\usepackage{float}

\begin{document}

\title{Machine Unlearning using Forgetting Neural Networks}

\author{\authorname{
Amartya Hatua\sup{1}\orcidAuthor{0000-0001-7868-0091}, 
Trung T. Nguyen\sup{2}\orcidAuthor{0000-0002-6695-2034}, 
Filip Cano\sup{3}\orcidAuthor{0000-0002-0783-904X}, and
Andrew H. Sung\sup{4}\orcidAuthor{0009-0005-0815-3102},
}
\affiliation{\sup{1}Fidelity Investments, Boston, USA}
\affiliation{\sup{2}Winona State University, Winona, USA}
\affiliation{\sup{3}Institute of Science and Technology Austria, Klosterneuburg, Austria}
\affiliation{\sup{4}Unversity of Southern Mississippi, USA}
\email{amartyahatua@gmail.com, trung.nguyen@winona.edu, filip.cano@ist.ac.at, andrew.sung@usm.edu}
}


\abstract{Modern computer systems store vast amounts of personal data, enabling advances in AI and ML but risking user privacy and trust. For privacy reasons, it is sometimes desired for an ML model to forget part of the data it was trained on. 
In this paper, we introduce a novel unlearning approach based on \emph{Forgetting Neural Networks} (FNNs), a neuroscience-inspired architecture that explicitly encodes forgetting through multiplicative decay factors. While FNNs had previously been studied as a theoretical construct, we provide the first concrete implementation and demonstrate their effectiveness for targeted unlearning. We propose several variants with per-neuron forgetting factors, including rank-based assignments guided by activation levels, and evaluate them on MNIST and Fashion-MNIST benchmarks. 
Our method systematically removes information associated with forget sets while preserving performance on retained data. Membership inference attacks confirm the effectiveness of FNN-based unlearning in erasing information about the training data from the neural network.
These results establish FNNs as a promising foundation for efficient and interpretable unlearning.}

\onecolumn \maketitle \normalsize \setcounter{footnote}{0} \vfill

\thispagestyle{plain}
\renewcommand{\thefootnote}{}
\footnotetext{Accepted at ICAART 2026 - 18th International Conference on Agents and Artificial Intelligence}
\renewcommand{\thefootnote}{\arabic{footnote}}

\section{Introduction}
\label{sec:intro}


Modern machine learning systems are increasingly trained on personal or sensitive data. Legal frameworks (e.g., right-to-erasure provisions) and organizational policies therefore require the ability to \emph{remove} the influence of specific data points after training~\cite{JULIUSSEN2023105885}. Beyond privacy, there are also performance-related motivations: as data evolves over time, previously relevant information may become obsolete or even misleading, and it is often desirable to efficiently eliminate such outdated influences from a model. Retraining from scratch on the retained data is the gold standard but is typically computationally prohibitive. \emph{Machine unlearning} addresses these challenges by aiming to efficiently transform a trained model so that its predictions (and internal representations) are indistinguishable from those of a model that had never seen the “forget” data, while maintaining utility on the remaining, up-to-date distribution.

Biological learning systems routinely balance acquisition and deletion of information via homeostatic and plasticity mechanisms such as synaptic scaling, long-term depression, and consolidation (e.g.\ synaptic stabilization, metaplasticity) \cite{Turrigiano2012,YgerGilson2015}. These processes suggest algorithmic primitives such as gating, decay, and homeostasis, which (i) make where and how much forgetting explicit, (ii) provide interpretable knobs that can be audited, and (iii) integrate naturally with standard optimization routines. Grounding unlearning in such neuro-inspired constructs can thus enhance \emph{trustworthiness} (by making failure modes and controls more transparent) and \emph{explainability} (by attributing changes to explicit forgetting factors rather than opaque fine-tuning) \cite{Ororbia2024survey}.

\paragraph{Forgetting Neural Networks.}
We study a family of neural architectures that include a forgetting term, called Forgetting Neural Networks (FNN).
FNNs were introduced in~\cite{cano2017theory} as fully connected or convolutional neural networks, with a per-layer forgetting factor that decays exponentially with time, inspired in the way that human brains experience forgetting.
In their original work, Cano-C{\'o}rdoba et al. propose this architecture as a sound way to transport neuroscience-inspired methods to the phenomenon of forgetting, and prove results linking the speed of forgetting with the decay of the parameters of the underlying neural network to their default value.

\paragraph{Machine Unlearning with FNNs.}
In this paper, we extend the original FNN architecture to a per-neuron forgetting factor, and propose a novel mechanism to use FNNs for targeted unlearning.
In a machine unlearning task, the input space is divided into two sets: the retain set and the forget set. 
The goal is to modify a neural network properly trained in the whole input space to perform poorly on the forget set, while retaining the performance of the original network on the retain set.
During an unlearning phase with our method, each factor multiplicatively attenuates either 
activation values and weights of each neuron, according to their forgetting parameter.
By targeting high forgetting parameters to the neurons most activated on datapoints in the forget set, we are able to steer the model towards erasing information associated with the forget set, while maintaining performance on the rest of the data.
We use membership inference attacks (MIAs) to measure the amount of information still present in the network about the forget set.
We propose an incremental algorithm that alternates phases of learning, unlearning, and MIA testing, which lets the user tune how much information the network forgets.

\paragraph{Experimental Evaluation.}
In our experimental evaluation, we study the effect of forgetting rate assignment, concluding that the most efficient is to assign forgetting rates to each neuron in accordance to its activation in the forget set: neurons that are most activated with instances of the forget set are the ones that receive a more aggressive forgetting effect.

Our evaluation protocol compares FNN-style unlearning against retraining and strong baselines for vision-based tasks. We report test accuracy and membership-inference attacks. 
We find that the improved FNNs achieve near-parity with retraining on retained utility while approaching chance-level membership detection on the forget set. 

\paragraph{Contribution.}
The main contributions of this paper are the following.
\begin{enumerate}
    \item We extend forgetting neural networks to per-neuron forgetting factors and provide the \emph{first implementation} of FNNs,
    providing an empirical demonstration of their theoretical properties.
    \item We introduce a \emph{novel method for machine unlearning} 
    using forgetting neural networks, and thus grounded in neuroscience-inspired forgetting mechanisms.
    \item We report on an  \emph{empirical validation}
    of our unlearning method on standard classification benchmarks for vision tasks (MNIST Digit and Fashion datasets).
\end{enumerate}

\paragraph{Outline of the paper}
Section~\ref{sec:prelim} formalizes machine unlearning, membership inference, and FNNs. Section~\ref{sec:method} presents our machine unlearning method using FNNs, and Section~\ref{sec:experiments} presents the results of our empirical evaluation.
We discuss strengths and limitations in Section~\ref{sec:discussion}, related work in Section~\ref{sec:relatedwork}, and provide concluding remarks and pointers to future work in Section~\ref{sec:conclusion}.

\section{Preliminaries}
\label{sec:prelim}

\subsection{The problem of machine unlearning}
Let $\mathcal{A}$ denote a (possibly randomized) training algorithm that maps a dataset $D$ to parameters $\theta=\mathcal{A}(D)$. The dataset decomposes into a retained subset $R$ and a forget subset $F$ with $D=R\cup F$ and $R\cap F=\emptyset$. Upon a forget request for $F$, an unlearning procedure $\mathcal{U}$ transforms $\theta$ into $\tilde{\theta}=\mathcal{U}(\theta, F)$ such that $\tilde{\theta}$ is close to $\theta^\star=\mathcal{A}(R)$, the model obtained by retraining on the retained data only. 
We evaluate closeness via two desiderata:
\begin{enumerate}
    \item \textbf{Utility on the retain set.}
    Performance on the retained distribution $\mathcal{D}_R$ is preserved.
    This can be modeled as the test accuracy remaining within a tolerance $\varepsilon_{\mathrm{util}}$ of $\theta^\star$.
    Since obtaining training $\theta^\star$ through fresh training can be expensive, it is usual to compare to the performance of $\theta$ as a proxy of the performance of $\theta^\star$.
    \item \textbf{Unlearning}. 
    The unlearned model $\tilde\theta$ should not contain information about the class membership of elements in the forget set $F$.
    This can be evaluated through the ineffectiveness of membership inference attacks (MIA).
\end{enumerate}

\subsection{Membership Inference Attacks}
\label{sec:prelimMIA}
Membership inference attacks (MIAs) test whether a sample $x$ belonged to a model’s training set by exploiting the different behavior that models often exhibit on seen versus unseen data. 
In a typical formulation, the adversary is given black-box access to a trained model $f_\theta$ and a datapoint $(x,y)$, and must decide whether $(x,y)\in D$ (a \emph{member}) or $(x,y)\notin D$ (a \emph{non-member}). 
Since training generally reduces the loss on samples it has seen, the distribution of losses or output confidences can leak membership information. 
Formally, let $L(f_\theta(x),y)$ denote the loss of model $f_\theta$ on input $x$ with label $y$. 
A simple but effective attack is the likelihood-ratio test that compares the distribution of losses on members versus non-members:
\[
\Lambda(x,y) \;=\; \frac{\Pr\!\left(L(f_\theta(x),y)\mid (x,y)\in D\right)}{\Pr\!\left(L(f_\theta(x),y)\mid (x,y)\notin D\right)}.
\]
If $\Lambda(x,y) > \nu$ for some threshold $\nu$, the attack predicts $(x,y)$ was in the training set. 

The effectiveness of a machine unlearning procedure can be measured by how well it neutralizes such attacks. 
In particular, for a perfectly unlearned model $\tilde{\theta}$, the loss distribution on the forget set $F$ should be indistinguishable from that on held-out test data. 
In this ideal case, any MIA behaves like random guessing, with success probability $0.5$. 
Deviations from $0.5$ indicate residual information about $F$ still encoded in the model.

\paragraph{Loss-based MIA.}
Following prior work~\cite{ShokriSSS17,pmlr-v139-choquette-choo21a}, the evaluation in this paper instantiates the likelihood-ratio test described above, using the forget set $F$ and a disjoint test set as proxies for members and non-members, respectively. 
Concretely, at each epoch the model’s loss is computed on both $F$ and the test set, and the empirical separation of these two distributions defines the attack’s success rate. 
This success rate is reported as the \emph{MIA score}. 
An MIA score close to $0.5$ indicates that unlearning was effective, whereas values significantly above $0.5$ imply that the model still reveals membership information about $F$. 

\subsection{Forgetting Neural Networks}

A Forgetting Neural Network (FNN), first introduced in~\cite{cano2017theory}, is a variant of the classical convolutional or fully connected neural networks in which trainable parameters are augmented with a multiplicative \emph{forgetting factor}. The motivation comes from Ebbinghaus’s classical forgetting curve~\cite{ebbinghaus1885gedachtnis,GAY2016147}, which characterizes the decay of memory retention in humans over time. 
In analogy to the neural processes of forgetting in humans, FNNs explicitly encode forgetting into the network architecture by introducing
per-layer forgetting factors in the form of a decay function $\varphi(t)$, where $t$ represents time.
Typically, $\varphi(t) = e^{-t/\tau}$, 
where the parameter $\tau$ is the forgetting rate,
defined as the time needed for the neuron to be dampened by a factor of $e^{-1} \approx 37\%$.
The weights and biases of the FNN are obtained by multiplying the original weights and biases by the corresponding forgetting factor, modulating their value as the forgetting process advances.
In contrast to standard networks, where information is implicitly forgotten through optimization dynamics when new data is processed in training batches, FNNs make the forgetting process explicit and thus, interpretable.

Formally, in the single-layer case, when the output of each neuron can be expressed as a standard perceptron $\Sigma_{[\theta_w,\ \theta_b]}(x) = \sigma(\langle \theta_w,\ x \rangle+\theta_b)$,
the parameters of the corresponding FNN are scaled by $\varphi(t)$, yielding a temporal decay of the learned function, as expressed in Equation~\ref{eq:single_forget_neuron}:
\begin{equation}
\label{eq:single_forget_neuron}
    \Sigma_{[\theta_w, \theta_b]}(x;t) = \sigma\big(\langle \theta_w\cdot \varphi(t), x  \rangle + \theta_b\cdot \varphi(t)\big).
\end{equation}
Extending this principle to multiple neurons and multiple layers, each forgetting neuron introduces a multiplicative factor $\varphi_k(t)$, whose effect propagates through the network.

\subsection{Activation Levels}
To better target forgetting of a concrete set of datapoints, we need some way to understand which neurons play the most important role in assigning a label to the datapoints to forget. 
The \emph{activation level} of a neuron for a given dataset $F$ is defined as the average magnitude of its post-activation output when the datapoints in $F$ are passed through the network. 
Formally, let $z_j(x)$ denote the pre-activation input of neuron $j$ for a datapoint $x$, and let $\sigma(\cdot)$ be the activation function of that neuron. 
Then the activation level of neuron $j$ with respect to dataset $F$ is

\begin{equation}
\label{eq:activation_level}
A_j(F) \;=\; \frac{1}{|F|}\sum_{x \in F} \left| \sigma\!\left(z_j(x)\right)\right|.
\end{equation}

Intuitively, $A_j(F)$ measures how strongly neuron $j$ responds on average when the forget dataset $F$ is presented. 
A higher value indicates that the neuron is more involved in representing or memorizing features of the forget set.

\section{Machine Unlearning using FNNs}
\label{sec:method}
\subsection{Forgetting Network Variants}
\paragraph{Fixed vs. Varying Forgetting Rate.}
In their original construction, forgetting neural networks implement the same forgetting factor for all neurons in the same layer. 
This is what we call \emph{fixed forgetting rate} networks (FFR).
We extend the original construction by allowing different forgetting factors for different neurons inside the same layer, to better account for forgetting targeted information.
We call these networks \emph{varying forgetting rate} networks (VFR).

\paragraph{Types of VFR Networks.}
Among networks with a varying forgetting rate, we study four different methods to assign a forgetting rate to each neuron.
In the first two of them, the forgetting rate is designed to target those neurons that produce large activation levels in the forget set. 
The other two, which can be thought of as baselines, introduce varying forgetting rates in a less principled manner.
\begin{enumerate}
    \item  \emph{Rank forget rate}: The neurons in a layer are sorted in increasing order according to their activation level in the forget set, $A_j(F)$. 
    The neuron in the $j$-th position (i.e., the neuron with rank $j$) is assigned a forgetting rate of $j$.
    Therefore, the higher the activation level, the lower the rank, making the forgetting for that neuron happen faster.
    \item \emph{Top $N$ forget rate}: 
    The top $N$ neurons according to their activation level in the forget set are given a forgetting rate $\tau=\tau_0$, 
    while the rest of the neurons are set to not forget (equivalently, are set with a forgetting rate of $\tau=\infty$).
    \item \emph{Ordered forget rate}: The neurons in a layer are sorted according to their original lexicographical order, and assigned a forgetting rate based on their position in this order. 
    In this case, the order is fixed at the initialization of the network, and does not depend on activation levels.
    \item \emph{Random forget rate}: The neurons are assigned a random forgetting rate, sampled from some distribution $\mathcal D$ over the real positive numbers.
\end{enumerate}

\subsection{Unlearning Method}
Once a variant of the forgetting neural network has been chosen, we implement unlearning through an iterative process consisting of alternating \emph{learning} and \emph{unlearning} bouts. 
The goal of this process is to balance two competing objectives: on the one hand, retaining accuracy on the retain set, and on the other, systematically erasing the influence of the forget set. 

During a learning bout, the network is trained on the retain data, which helps the model maintain performance on non-sensitive samples and counteracts the loss of generalization introduced by forgetting. 
During an unlearning bout, the forget set is presented to the network, and the forgetting layers apply the decay function $\varphi(t) = e^{-t/\tau}$ to progressively weaken the neurons most responsible for encoding the forget data. 
This pairing of learning and unlearning ensures that forgetting is targeted at sensitive information, while performance on the remaining data is preserved. 

The process is repeated for $n$ turns, each consisting of a learning bout followed by an unlearning bout. 
Inside each learning bout

\section{Experimental Evaluation}
\label{sec:experiments}

In this section, we report the results of our empirical evaluation of performing machine unlearning with our FNN-based method, as described in Section~\ref{sec:method}.
The main goal of our experiments is to have a proof of concept showing the effectiveness of the unlearning algorithm, 
to show that varying forgetting rates induce better forgetting results than classical (fixed) ones, and to determine which one of the strategies to choose the variation in forgetting rates gives best results.


\begin{algorithm}[t]
\caption{Machine Unlearning using FNNs}\label{alg:unlearning}
\KwIn{training\_data, testing\_data, retain\_data, forget\_data}

Train model once on \textit{training\_data}\;

\For{$turn \gets 1$ \KwTo $\mathtt{N\_turns}$}{
    \tcc{Learning bout}
    \For{$\mathtt{epoch} \gets 1$ \KwTo $\mathtt{training\_epochs}$}{
      \textbf{I)} Train model on \textit{retain\_data}\;
      \textbf{II)} Evaluate accuracy on \textit{testing\_data}\;
      \textbf{III)} Evaluate MIA on \textit{forget\_data}\;
    }
    
    \tcc{Unlearning bout}
    \For{$\mathtt{epoch} \gets 1$ \KwTo $\mathtt{forget\_epochs}$}{
      \textbf{I)} Present \textit{forget\_data} and apply forgetting functions $\varphi(t=\mathtt{epoch})$\;
      \textbf{II)} Evaluate accuracy on \textit{testing\_data}\;
      \textbf{III)} Evaluate MIA on \textit{forget\_data}\;
    }
}
\end{algorithm}

\subsection{Experimental Setup}

\paragraph{Datasets.}
The MNIST handwritten digit recognition (HDR) dataset~\cite{deng2012mnist} has long been one of the standard benchmarks for vision-based ML classification problems. 
Following concerns that it may be too easy as a problem, Xiao at al.\ introduced the MNIST fashion dataset~\cite{xiao2017fashion}, as an alternative with the same format and a more challenging classification problem. 
We use both datasets in our experiments.

\paragraph{Neural Network.}
For both datasets, we employ the same neural network architecture. It consists of two convolutional layers, each followed by ReLU activation and max pooling, and two fully connected layers. A final softmax layer produces the classification scores for each class.

\begin{figure}[t]
\centering
\includegraphics[width=0.95\linewidth, height=4cm]{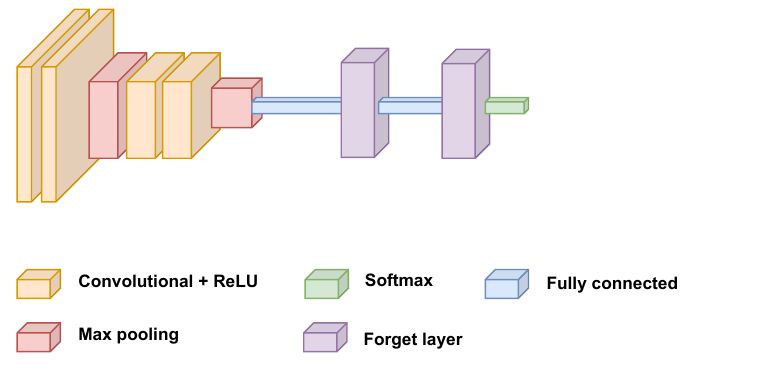}
\caption{Diagram of the implemented model. Built with the help of~\cite{FNN_Diagrams}.}
\label{FNN_CNN}
\par\bigskip
\end{figure}

We apply forgetting factors to the fully connected layers. 
In our evaluation, we experiment with applying forgetting to both fully connected layers, and only to the outermost one.
The architecture, comprising approximately one million trainable parameters, is illustrated in Figure~\ref{FNN_CNN}.

We denote experiments with forgetting in both fully connected layers as (2-L), and experiments where only the $FL2$ layer has active forgetting as (1-L).

\paragraph{Initial Training.}
Before applying any forgetting, we train the neural network on the full training split of each dataset using the Adam optimizer~\cite{adamKingmaB14} with a batch size of 64. This results in a test accuracy of approximately $95\%$.

\subsection{Accuracy of the Unleared Models}
In our first set of experiments, we track the accuracy of the neural network as Algorithm~\ref{alg:unlearning} progresses, for different types of forgetting schemes.
As algorithm parameters, we use $\mathtt{training\_epochs}=2$ and $\mathtt{forget\_epochs}=4$.

\begin{figure}[t]
\centering
    \begin{subfigure}{\linewidth}
        \centering 
        \includegraphics[width=\linewidth, height=5.5cm]{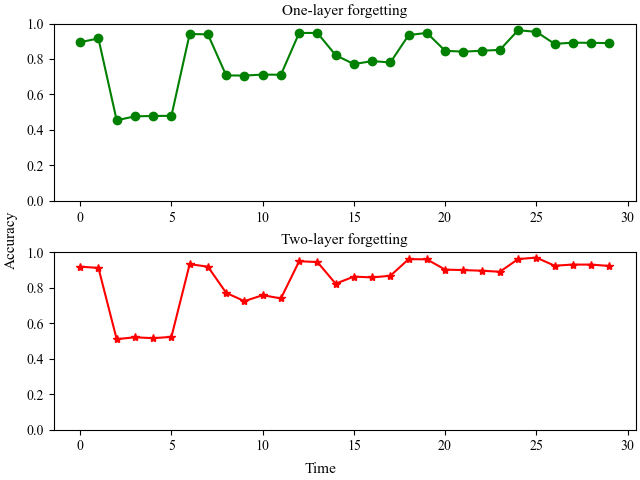}
        \caption{FFR FNN for MNIST fashion}
		\label{FASHION_fixed_rank_two_layer_plot}
    \end{subfigure}
    \begin{subfigure}{\linewidth}
        \centering 
        \includegraphics[width=\linewidth, height=5.5cm]{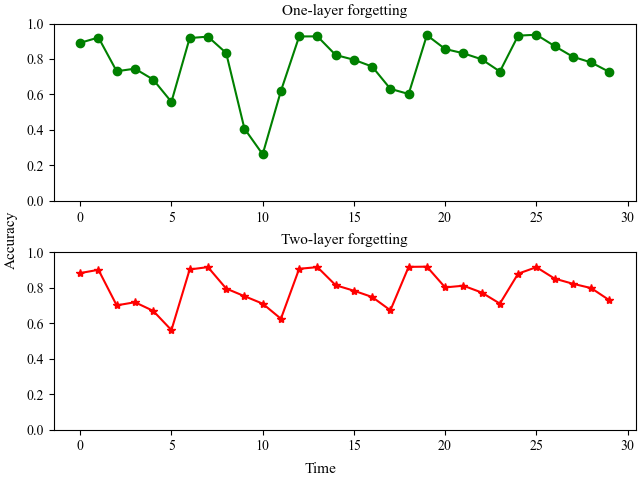}
        \caption{FFR FNN for MNIST HDR}
		\label{DIGIT_fixed_rank_two_layer_plot}
    \end{subfigure}
\caption{Learning-unlearning curve for FFR FNN }
\label{FNN_Deterministic}
\end{figure}

In Figure~\ref{FNN_Deterministic}, 
we show the learning-unlearning graph for fixed forgetting rate (FFR) models for both datasets.
We use a fixed forgetting rate of $\tau = 2$.

\begin{figure}[t]
\centering
    \begin{subfigure}{\linewidth}
        \centering 
        \includegraphics[width=\linewidth]{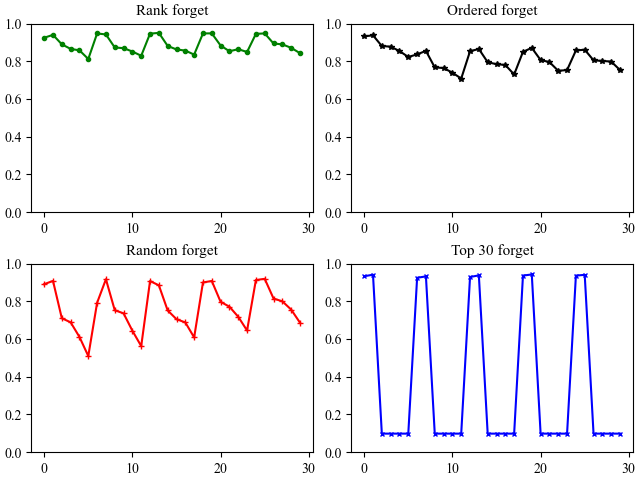}
        \caption{One forgetting layer (1-L)}
		\label{DIGIT_variable_epoch_one_layer_plot}
    \end{subfigure}
    \begin{subfigure}{\linewidth}
        \centering 
        \includegraphics[width=\linewidth]{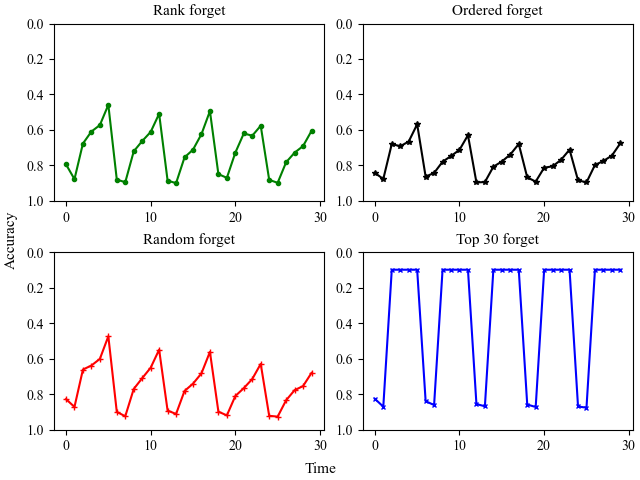}
        \caption{Two forgetting layers (2-L)}
		\label{DIGIT_variable_epoch_two_layers_plot}
    \end{subfigure}
\caption{Learning-unlearning curve for VFR FNN on the MNIST HDR dataset.}
\label{fig:VFR_accuracy_MNIST}
\end{figure}

\begin{figure}[h]
\centering
    \begin{subfigure}{\linewidth}
        \centering 
        \includegraphics[width=\linewidth]{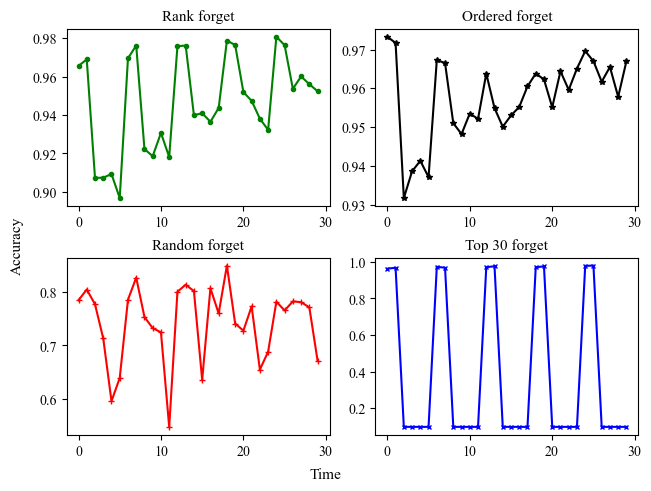}
        \caption{One forgetting layer (1-L)}
		\label{FASHION_variable_epoch_last_layer_plot}
    \end{subfigure}
    \begin{subfigure}{\linewidth}
        \centering 
        \includegraphics[width=\linewidth]{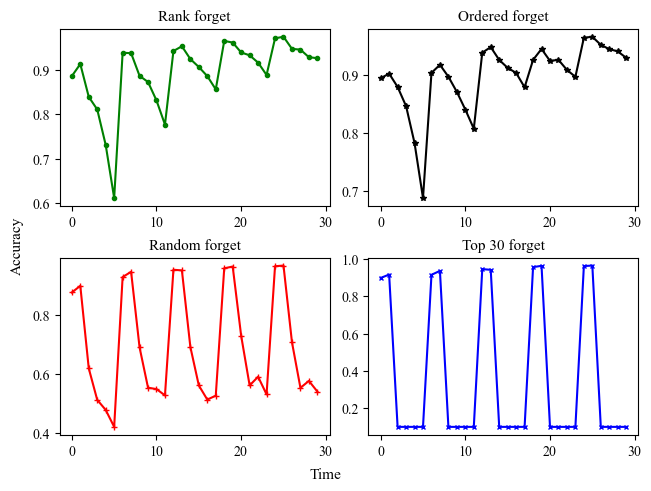}
        \caption{Two forgetting layers (2-L)}
		\label{FASHION_variable_epoch_two_layers_plot}
    \end{subfigure}
\caption{Learning-unlearning curves for VFR FNN on the MNIST Fashion dataset.}
\label{fig:VFR_accuracy_fashion}
\end{figure}

In Figures~\ref{fig:VFR_accuracy_MNIST} and~\ref{fig:VFR_accuracy_fashion}, 
we show the learning-unlearning curves for the different types of varying forgetting rates (VFR).
We use $N=30$ for the Top-$N$ rate.
To keep a fair comparison among the different forgetting schemes with varying rate and also with the fixed rate scheme, in all methods that apply a single forgetting rate to some of the neurons (i.e., fixed, random, and Top 30), we use the same forgetting factor of $\tau=2$. 

We can already observe the general pattern that the accuracy increases with learning phases and decreases with unlearning phases, which was to be expected. We see more consistent  results for the HDR dataset than the Fashion dataset, which may stem from the fact that Fashion classification is a more challenging task, so performance can be lost quickly. The graphs also show the tendency, however, that subsequent learning-unlearning turns help increase performance.
Among the different strategies for varying forgetting rates, \emph{random} shows the most inconsistent results, which was to be expected.
\emph{Rank} shows the best performance results, together with \emph{ordered} in some cases. 
Rank is the most principled way of defining forgetting rates. We think that \emph{ordered} may be the one that benefits the most from having several learning-unlearning turns, because the network can adapt the neurons responsible to learn the retain set at each turn, as the neurons with high forgetting rate are kept fixed through the turns.
Top 30 shows the steepest unlearning, with a big portion of the accuracy being lost already in the first forgetting epoch. This phenomenon suggests that it may be too aggressive as a forgetting strategy.

We conclude that rank forgetting is the strategy that shows best results among the VRF strategies, so in the next phase of the empirical evaluation we will only use Rank as a VFR scheme.

\subsection{Membership Inference Attacks}

\paragraph{Setup.}
As discussed in Section~\ref{sec:prelimMIA}, membership inference attacks (MIAs) are a standard tool for evaluating unlearning performance. 
In an MIA, an adversary attempts to determine whether a given datapoint was part of the training set. 
The resulting MIA score ranges from $0$ to $1$, where a score of $0$ corresponds to high confidence that the datapoint was not included in training, and a score of $1$ corresponds to high confidence that it was included. 
For machine unlearning, a score closer to $0.5$ indicates better performance, as it reflects indistinguishability between training and non-training data. 
We implement MIAs using the Adversarial Robustness Toolbox (ART) \cite{art2018}.
Following standard practices in machine unlearning\footnote{See, for example, the NeurIPS 2023 Machine Unlearning Challenge, \\ \url{https://unlearning-challenge.github.io/}.}, 
as a baseline we use the \emph{retrain-from-scratch} method, which consists of training the neural network from scratch on the fraction of the train dataset that contains no points from the forget set.

\begin{figure}
\centering
    \begin{subfigure}{\linewidth}
        \centering 
        \includegraphics[width=\linewidth]{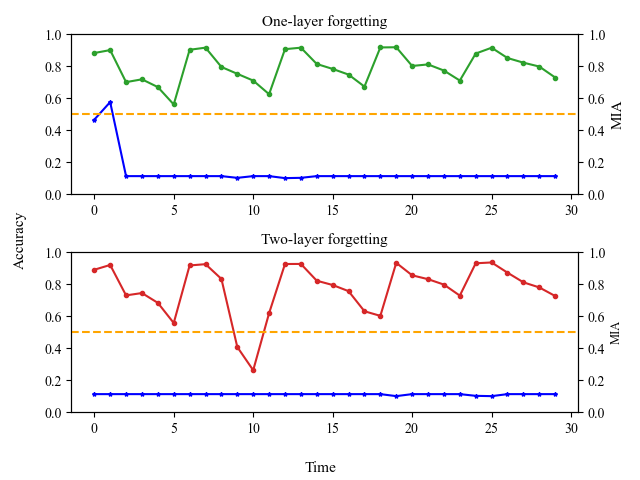}
        \caption{MNIST HDR}
		\label{DIGIT_fixed_rank_Accuracy_MIA}
    \end{subfigure}
    \begin{subfigure}{\linewidth}
        \centering 
        \includegraphics[width=\linewidth]{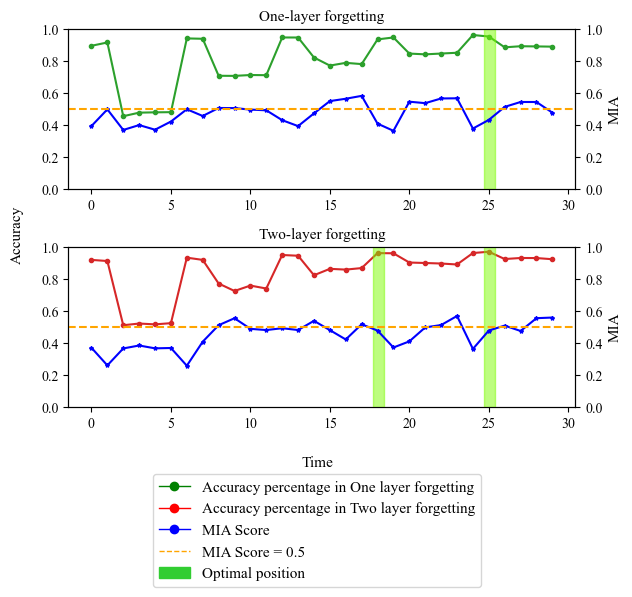}
        \caption{MNIST Fashion}
		\label{FASHION_fixed_rank_Accuracy_MIA}
    \end{subfigure}
\caption{Learning-unlearning curve \& MIA score for fixed forgetting rate networks.}
\label{fig:MIA-FFR}
\end{figure}

\begin{figure}
\centering
    \begin{subfigure}{\linewidth}
        \centering 
        \includegraphics[width=\linewidth]{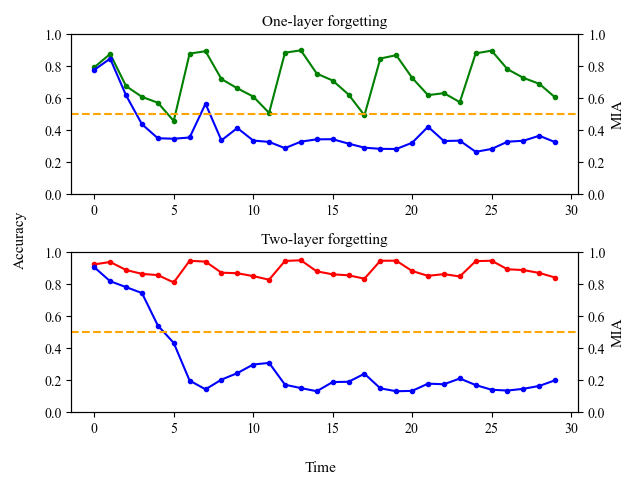}
        \caption{MNIST HDR}
		\label{DIGIT_variable_epoch_two_layers_Accuracy_MIA}
    \end{subfigure}
    \begin{subfigure}{\linewidth}
        \centering 
        \includegraphics[width=\linewidth]{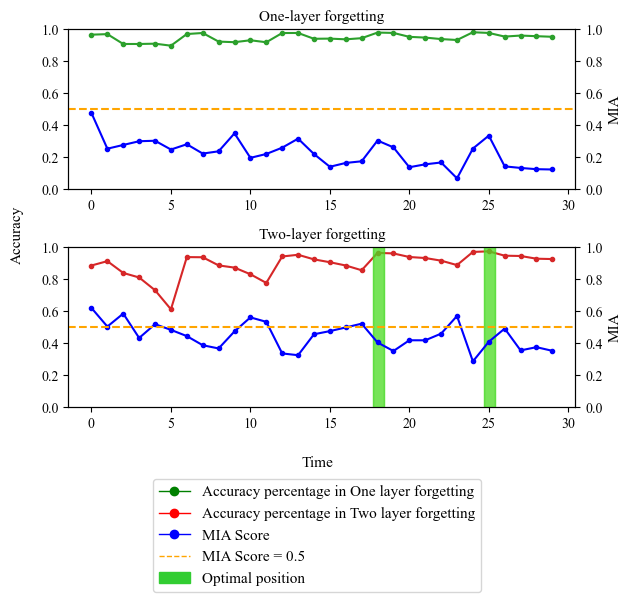}
        \caption{MNIST fashion}
		\label{FASHION_variable_epoch_two_layers_Accuracy_MIA}
    \end{subfigure}
\caption{Learning-unlearning curve \& MIA score for Rank VFR forgetting networks.}
\label{fig:MIA-VFR}
\end{figure}

\paragraph{Unlearning Performance of FNNs.}
In Figures~\ref{fig:MIA-FFR} and~\ref{fig:MIA-VFR}, we show the learning-unlearning curves, together with the MIA scores, for our different forgetting schemes (FFR and Rank VFR, 1-L and 2-L).
In general, we observe an improvement as the learn-unlearn turns progress. 
As our ``final forgetting objective'', we consider that a neural network that retains accuracy over $95\%$ and an MIA score of $0.5 \pm 0.1$ has achieved ``optimal forgetting''. 
These ``optimal forgetting'' timesteps are marked in green in Figures~\ref{fig:MIA-FFR} and~\ref{fig:MIA-VFR}. 

\begin{figure}[t]
\centering
\includegraphics[width=\linewidth]{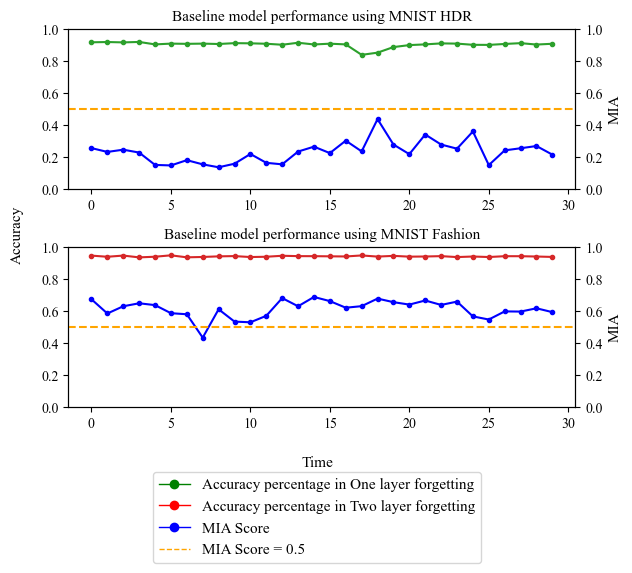}
\caption{Baseline Result}
\label{baseline_result}
\end{figure}

\paragraph{Comparison to Baseline.}
As mentioned before, we use the \emph{retrain-from-scratch} method as our natural baseline for unlearning. 
The results of this baseline method are reported in Figure~\ref{baseline_result}.
We can observe that the baseline method achieves a good performance in terms of both accuracy and MIA from the beginning (i.e., it does not benefit from subsequent turns, as there is no learning-unlearning loop, only the same learning task).
We also observe that, although the model shows high accuracy and good MIA scores, at no point do they achieve what we defined as ``optimal forgetting''. 
This phenomenon of over-shooting the unlearning is a documented issue in machine unlearning literature, known as ``over-forgetting''. We discuss it further in the following section.

\section{Discussion}
\label{sec:discussion}

\begin{figure}[t]
    \centering
    \includegraphics[width=0.65\linewidth]{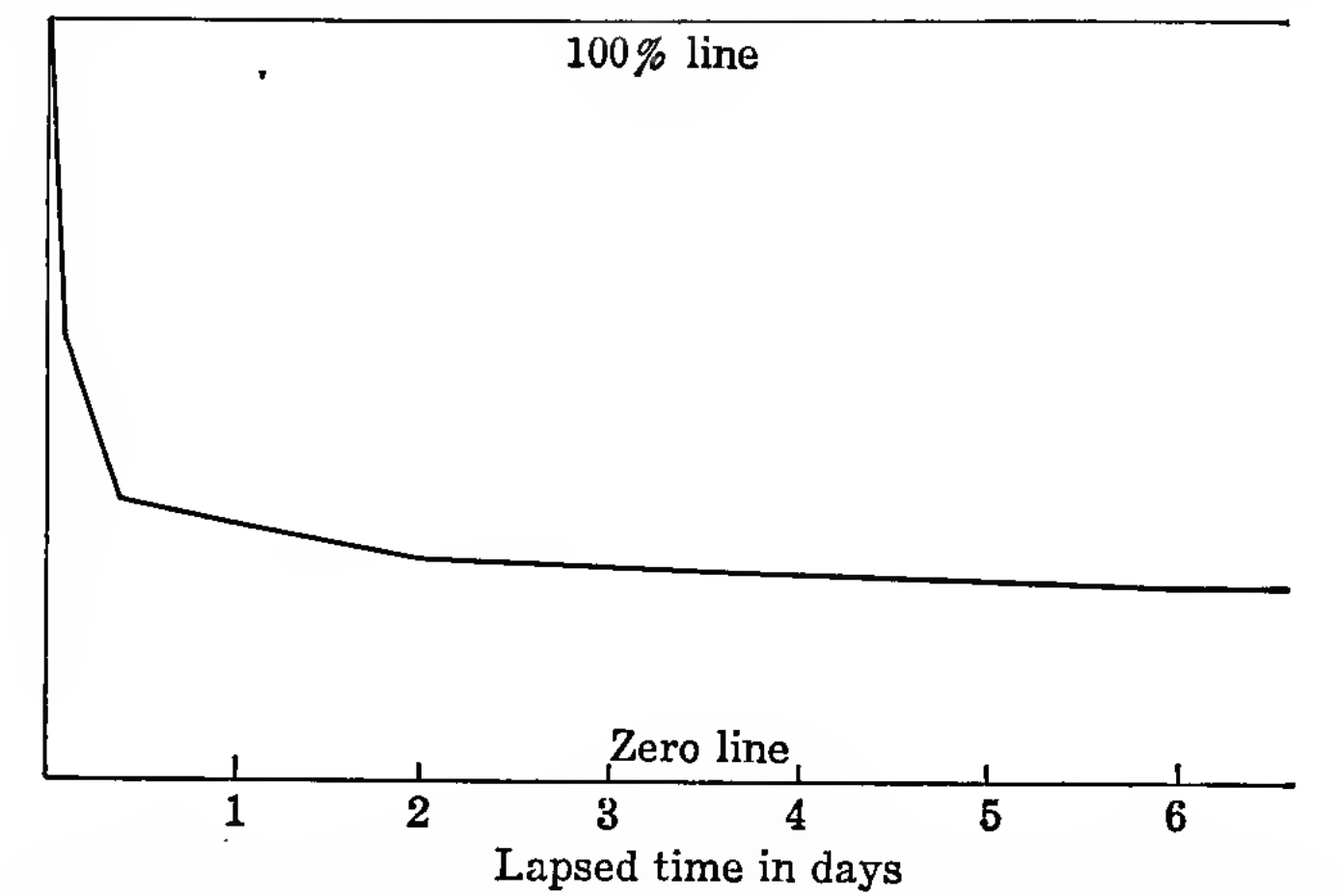}
    \caption{Early depiction of the Ebbinghaus forgetting curve. Image taken from~\cite[Fig. 53]{psychology1921}.}
    \label{fig:ebb_origina_graph}
\end{figure}

\paragraph{Neuroscience-Inspired Methods in ML.}
The Ebbinghaus forgetting curve is one of the earliest documented phenomena in the scientific study of memory, demonstrating that forgetting follows a systematic temporal pattern. Ebbinghaus’s work~\cite{ebbinghaus1885gedachtnis} marked a transition from philosophical speculation to experimental measurement, laying the foundation for modern cognitive modeling. See Figure~\ref{fig:ebb_origina_graph} for an early graphical depiction of this curve, using Ebbinhaus original data.
Our method highlights the value of drawing inspiration from cognitive science and neuroscience in the design of ML algorithms. By embedding explicit forgetting factors into the network architecture, the process of unlearning becomes more transparent: information is weakened in controlled, interpretable ways rather than hidden in the opaque optimization dynamics of fine-tuning or retraining. 
This allows practitioners to reason about where forgetting occurs and how strongly it impacts particular neurons. 
Moreover, the analogy with human memory processes (e.g., Ebbinghaus’ forgetting curve) provides an intuitive conceptual grounding, which can increase trust and explainability. At the same time, this biological analogy should be seen as a useful heuristic rather than a literal model: the simplifications made in FNNs only partially capture the complexity of human forgetting.

\paragraph{Other VFR methods.}
Among the proposed strategies, the rank-based varying forgetting rate (VFR) proved most effective in balancing utility and forgetting. This suggests that activation levels provide a reasonable proxy for which neurons encode forget-set information. However, the performance gain over the fixed forgetting rate (FFR) baseline was not dramatic, and other principled ways of assigning forgetting factors remain unexplored. 
Additionally, we only study forgetting on the fully connected layers, because it is there where more conceptual-level information is encoded, and it would be interesting to see what is forgotten when dampening the convolutional layers.


\paragraph{The over-forgetting phenomenon.} 
Interestingly, we observe that Rank VFR sometimes achieves both high accuracy and near-random MIA scores, while the retrain-from-scratch baseline does not meet this joint condition. Since retraining is the gold standard for exact removal, this discrepancy highlights a conceptual difference: retraining corresponds to \emph{neutral forgetting} (as if the samples had never been seen), whereas forgetting neural networks implement an \emph{active forgetting} mechanism that deliberately dampens the signal of forget samples. In this sense, FNNs behave less like retraining and more like human memory suppression, where unwanted experiences are not merely absent but actively erased together with their associative traces.
This phenomenon, known in the literature as \emph{over-forgetting}, has been documented in many machine unlearning approaches~\cite{zhao2024makes,he2025towards}.
Such behavior reflects the fact that many methods, including ours, implement active suppression mechanisms rather than neutral removal. 
While this can reduce the effectiveness of membership inference attacks, a lower MIA score does not necessarily mean a ``better'' unlearning outcome, as it may instead indicate that the model has entered an over-forgetting regime.


\paragraph{Comparison to State of the Art.}
In machine unlearning, retraining from scratch on the retained dataset naturally serves as the gold standard, since it guarantees removal of the forget set’s influence. For this reason, comparison to other state-of-the-art methods is less critical than in many other ML domains: the key benchmark is how closely an unlearning method approximates retraining in terms of both utility and privacy. Our results show that FNN-based unlearning approaches this gold standard, offering similar accuracy and protection against membership inference.

\paragraph{Efficiency and Guarantees.}
While our method has the advantage of being principled and interpretable, 
optimization-based methods provide faster erasure with fewer epochs, 
and other methods incorporate guarantees of exact equivalence to retraining under certain assumptions. In contrast, FNNs currently lack formal guarantees of correctness: our experiments show promising empirical forgetting, but we cannot theoretically certify that the influence of the forget set is fully removed, other than the limit for $t\to\infty$, where the whole network converges to its default value, as already pointed out in~\cite{cano2018theoretical}.

\section{Related Work}
\label{sec:relatedwork}

There is a vast body of literature on machine unlearning, reflecting its importance for privacy, fairness, and accountability in modern AI systems~\cite{bourtoule2021machine,nguyen2022survey,qu2023learn,zhang2023review}. 
Unlearning methods can be broadly divided into approaches for neural networks and those targeting more traditional machine learning models.

\paragraph{Unlearning with Neural Networks.}
Given the widespread adoption of deep learning models, unlearning methods specifically designed for neural networks have gathered considerable attention. 
Early work by Golatkar et al.\ proposed to estimate the contribution of forget samples to the weights of a trained DNN and remove them via perturbation~\cite{DBLP:journals/corr/abs-1911-04933}, later improving their approach through linear approximations that cast the unlearning task as a quadratic optimization problem~\cite{golatkar2021mixedprivacy}. 
Kim et al.\ developed a bias-prediction network to selectively unlearn biased information from learned embeddings~\cite{Kim_2019_CVPR}. 
More recently, Chundawat et al.\ introduced a student--teacher framework together with the Zero Retrain Forgetting (ZRF) metric to guide knowledge transfer while removing the influence of forget samples. 
Hoang et al.\ proposed Projected Gradient Unlearning (PGU), which preserves knowledge of the retain set by projecting updates into the orthogonal complement of its gradient subspace~\cite{Hoang_2024_WACV}. 

In parallel, several works have focused on \emph{certified unlearning}, where the aim is to provide formal guarantees that forgetting is equivalent (or close) to retraining from scratch. 
Neel et al.\ introduced the Descent-to-Delete framework for convex objectives~\cite{neel2020descent}. 
Sekhari et al.\ formalized certified removal in convex optimization~\cite{sekhari2021remember}. 
Thudi et al.\ extended these ideas to stochastic gradient descent by unrolling updates~\cite{thudi2022unrolling}. 
Izzo et al.\ analyzed approximate unlearning in deep networks and proposed influence-based algorithms~\cite{izzo2021approximate}. 
Guo et al.\ provided certificates of removal applicable to a broad class of models~\cite{guo2020certified}, and
subsequent work has improved efficiency~\cite{tarun2023fast}.

\paragraph{Unlearning with Traditional ML.}
Before the focus shifted to deep networks, unlearning was studied in more classical settings. 
For Support Vector Machines (SVMs), decremental learning methods were proposed to update models efficiently when samples are removed~\cite{Duan2007DecrementalLA,5484614}. 
Cao and Yang~\cite{cao2015towards} introduced a general framework that transforms certain learning algorithms into summation forms, enabling the removal of samples by updating corresponding statistics. 
Brophy and Lowd later developed the Data-Removal-Enabled (DARE) forests for random forests, which cache node statistics to permit efficient unlearning~\cite{brophy2021machine}. 
Other directions explored Bayesian models~\cite{nguyen2020variational} and clustering models, such as K-means, where Ginart et al.\ introduced the first sublinear-time exact unlearning algorithm~\cite{ginart2019making}. 

\paragraph{Connections to Privacy Attacks.}
Finally, there is a close connection between unlearning and membership inference attacks (MIAs)~\cite{ShokriSSS17,yeom2018privacy,pmlr-v139-choquette-choo21a}. 
Since the goal of unlearning is to remove traces of forget samples from a model, effectiveness is often evaluated by the inability of MIAs to distinguish forget samples from unseen data. 
This connection has motivated the use of MIAs as an evaluation standard across many unlearning studies, including ours.

\paragraph{Neuroscience-based approaches to ML.}
The ideas presented in this paper are part of a broader trend in AI research to draw inspiration on neuroscience mechanisms.
For example, context-dependent gating—where only sparse, mostly nonoverlapping subsets of hidden units activate per task—was proposed as a neuroscience-inspired mechanism to reduce interference in continual learning \cite{MasseGrantFreedman2018}. 
Synaptic consolidation or metaplastic control (i.e.\ regulating plasticity based on historical importance) similarly provides a regulated forgetting mechanism \cite{Jedlicka2022,TheoriesSynapticMemoryConsolidation2024}. In spiking neural models, adaptive synaptic decay (e.g.\ Adaptive Synaptic Plasticity) explicitly “leaks” weights unless reinforced by new evidence \cite{PandaAllredRamanathanRoy2017}.  
Finally, offline replay or sleep-like reactivation (as a form of consolidation) has been empirically observed to mitigate forgetting in artificial networks \cite{Tadros2022,Golden2022}. Together these illustrate that neuroscience-inspired forgetting and adaptation mechanisms are viable, interpretable, and potentially robust tools for trustworthy unlearning in AI systems.

\section{Conclusion and Future Work}
\label{sec:conclusion}
We presented the first application of Forgetting Neural Networks (FNNs) to the problem of machine unlearning. By introducing per-neuron forgetting factors and targeted unlearning strategies, our method translates neuroscience-inspired forgetting mechanisms into practical ML tools. Our experiments on MNIST and Fashion-MNIST showed that FNN-based unlearning preserves utility on the retain set while effectively neutralizing membership inference attacks, with performance approaching that of the retraining gold standard. Among the studied configurations, the rank-based varying forgetting rate appears to be the most reliable strategy.

Future work should address several limitations. First, our evaluation was restricted to small-scale vision benchmarks; scaling FNN unlearning to more complex datasets and modern architectures (e.g., transformers) is a natural next step. 
Second, the current approach handles a single forget request; extending it to sequential and overlapping requests would broaden its applicability. 
Finally, the explicit structure of FNN forgetting factors suggests opportunities beyond privacy: for example, adapting the mechanism to enforce fairness constraints, remove spurious correlations, or provide interpretable guarantees of compliance with right-to-erasure requirements.


\section*{Acknowledgements}
This work has been supported by the European Research Council under Grant No.: ERC-2020-AdG 101020093.
We thank Brian Subirana for the original ideation of the concept of forgetting neural networks and his belief in their potential.

\bibliographystyle{apalike}

\bibliography{references}

\end{document}